\documentclass[11pt]{article}
\usepackage{acl2016}
\usepackage{times}
\usepackage{url}
\usepackage{latexsym}
\usepackage[utf8]{inputenc}
\usepackage[T2A,T1]{fontenc}
\usepackage[russian,english]{babel} 
\usepackage[labelsep=period,labelfont=bf,figurename={Figure},tablename={Table},figurewithin=none]{caption}
\usepackage{graphicx}
\usepackage{booktabs}
\usepackage{multicol}

\aclfinalcopy

\title{Redefining part-of-speech classes with distributional semantic models}

\author{Andrey Kutuzov \\
  Department of Informatics\\
  University of Oslo\\
  {\tt andreku@ifi.uio.no} \\ \And
  Erik Velldal \\
  Department of Informatics\\
  University of Oslo\\
  {\tt erikve@ifi.uio.no} \\ \And
  Lilja Øvrelid \\
  Department of Informatics\\
  University of Oslo\\
  {\tt liljao@ifi.uio.no}}

\begin{document}
\maketitle
 
 
\begin{abstract}
This paper studies how word embeddings trained on the British National Corpus interact with part of speech boundaries. Our work targets the Universal PoS tag set, which is currently actively being used for annotation of a range of languages. We experiment with training classifiers for predicting PoS tags for words based on their embeddings. The results show that the information about PoS affiliation contained in the distributional vectors allows us to discover groups of words with distributional patterns that differ from other words of the same part of speech. 

This data often reveals hidden inconsistencies of the annotation process or guidelines. At the same time, it supports the notion of `soft' or `graded' part of speech affiliations. Finally, we show that information about PoS is distributed among dozens of vector components, not limited to only one or two features.
\end{abstract}

\section{Introduction}
Parts of speech (PoS) are useful abstractions, but still abstractions. Boundaries between them in natural languages are flexible. Sometimes, large open classes of words are situated on the verge between several parts of speech: for example, participles in English are in many respects both verbs and adjectives. In other cases, closed word classes `intersect', e.g., it is often difficult to tell a determiner from a possessive pronoun. As \newcite{houston1985continuity} puts it, `\textit{Grammatical categories exist along a continuum which does not exhibit sharp boundaries between the categories}'.

When annotating natural language texts for parts of speech, the choice of a PoS tag in many ways depends on the human annotators themselves, but also on the quality of linguistic conventions behind the division into different word classes. That is why there have been several attempts to refine the definitions of parts of speech and to make them more empirically grounded, based on corpora of real texts: see, among others, the seminal work of \newcite{biber1999longman}. The aim of such attempts is to identify clusters of words occurring naturally and corresponding to what we usually call `parts of speech'. One of the main distance metrics that can be used in detecting such clusters is a distance between distributional features of words (their contexts in a reference training corpus).

In this paper, we test this approach using predictive models developed in the field of distributional semantics. Recent achievements in training distributional models of language using machine learning allow for robust representations of natural language semantics created in a completely unsupervised way, using only large corpora of raw text. Relations between dense word vectors (embeddings) in the resulting vector space are as a rule used for semantic purposes. But can they be employed to discover something new about grammar and syntax, particularly parts of speech? Do learned embeddings help here? Below we show that such models do contain a lot of interesting data related to PoS classes.

The rest of the paper is organized as follows. In Section \ref{sec:related} we briefly cover the previous work on the subject of parts of speech and distributional models. Section \ref{sec:pos} describes data processing and the training of a PoS predictor based on word embeddings. In Section \ref{sec:crowd} errors of this predictor are analyzed and insights gained from them described. Section \ref{sec:predict} introduces an attempt to build a full-fledged PoS tagger within the same approach. It also analyzes the correspondence between particular word embedding components and PoS affiliation, before we conclude in Section \ref{sec:conclusion}.

\section{Related work}\label{sec:related}


Traditionally, 3 types of criteria are used to distinguish different parts of speech: \textit{formal} (or morphological), \textit{syntactic} (or distributional) and \textit{semantic} \cite{aarts2008handbook}. Arguably, syntactic and semantic criteria are not very different from each other, if one follows the famous distributional hypothesis stating that meaning is determined by context \cite{firth1957synopsis}. Below we show that unsupervised distributional semantic models contain data related to parts of speech.

For several years already it has been known that some information about morphological word classes is indeed stored in distributional models.  Words belonging to different parts of speech possess different contexts: in English, articles are typically followed by nouns, verbs are typically accompanied by adverbs and so on. It means that during the training stage, words of one PoS should theoretically cluster together or at least their embeddings should retain some similarity allowing for their  separation from words belonging to other parts of speech. Recently, among others, \newcite{tsuboi:2014} and \newcite{plank2016multilingual} have demonstrated how word embeddings can improve supervised PoS-tagging.

\newcite{mikolov2013linguistic} showed that there also exist regular relations between words from different classes: the vector of `\textit{Brazil}'is related to `\textit{Brazilian}' in the same way as `\textit{England}' is related to `\textit{English}' and so on. Later, \newcite{liu2016part} demonstrated how words of the same part of speech cluster into distinct groups in a distributional model, and \newcite{tsvetkov2015} proved that dimensions of distributional models are correlated with different linguistic features, releasing an evaluation dataset based on this.

Various types of distributional information has also played an important role in previous work done on the related problem of unsupervised PoS acquisition. As discussed in \newcite{christodoulopoulos2010two}, we can separate at least three main directions within this line of work: \emph{Disambiguation} approaches \cite{merialdo1994multi,toutanova2007bayesian,ravi2009minimized} that start out from a dictionary providing possible tags for different words; \emph{prototype-driven} approaches \cite{haghighi2006prototype,christodoulopoulos2010two} based on a small number of prototypical examples for each PoS; \emph{induction} approaches that are completely unsupervised and make no use of prior knowledge. This is also the main focus of the comparative survey provided by \cite{christodoulopoulos2010two}.


Work on PoS induction has a long history -- including the use of distributional methods -- going back at least to \newcite{schutze1995distributional}, and recent work has demonstrated that word embeddings can be useful for this task as well \cite{yatbaz2012learning,lin2015unsupervised,ling2015not}.

In terms of positioning this study relative to previous work, it falls somewhere in between the distinctions made above. It is perhaps closest to disambiguation approaches, but it is not unsupervised given that we make use of existing tag annotations when training our embeddings and predictors. The goal is also different; rather than performing PoS acquisition or tagging for its own sake, the main focus here is on analyzing the boundaries of different PoS classes. In Section \ref{sec:predict}, this analysis is complemented by experiments with using word embeddings for PoS prediction on unlabeled data, and here our approach can perhaps be seen as related to previous so-called prototype-driven approaches, but in these experiments we also make use of labeled data when defining our prototypes.

It seems clear that one can infer data about PoS classes of words from distributional models in general, including embedding models. As a next step then, these models could also prove useful for deeper analysis of part of speech boundaries, leading to discovery of separate words or whole classes that tend to behave in non-typical ways. Discovering such cases is one possible way to improve the performance of existing automatic PoS taggers \cite{manning2011part}. These `outliers' may signal the necessity to revise the annotation strategy or classification system in general. Section \ref{sec:pos} describes the process of constructing typical PoS clusters and detecting words that belong to a cluster different from their traditional annotation. 

\section{PoS clusters in distributional models} \label{sec:pos}

Our hypothesis is that for the majority of words their parts of speech can be inferred from their embeddings in a distributional model. This inference can be considered a classification problem: we are to train an algorithm that takes a word vector as input and outputs its part of speech. If the word embeddings do contain PoS-related data, the properly trained classifier will correctly predict PoS tags for the majority of words: it means that these lexical entities conform to a dominant distributional pattern of their part of speech class. At the same time, the words for which the classifier outputs \textit{incorrect} predictions, are expected to be `outliers', with distributional patterns different from other words in the same class. These cases are the points of linguistic interest, and in the rest of the paper we mostly concentrate on them.

To test the initial hypothesis, we used the XML Edition of British National Corpus (BNC), a balanced and representative corpus of English language of about 98 million word tokens in size.  As stated in the corpus documentation, `\textit{it was} [PoS-]\textit{tagged automatically, using the CLAWS4 automatic tagger developed by Roger Garside at Lancaster, and a second program, known as Template Tagger, developed by Mike Pacey and Steve Fligelstone}' \cite{burnard_bnc}. The corpus authors report a precision of 0.96 and recall of 0.99 for their tools, based on a manually checked sample. For this research, it is important that BNC is an established and well-studied corpus of English with PoS-tags and lemmas assigned to all words.

We produced a version of BNC where all the words were replaced with their lemmas and PoS-tags converted into the Universal Part-of-Speech Tagset \cite{petrov2012universal}\footnote{We used the latest version of the tagset available at \url{http://universaldependencies.org}}. Thus, each token was represented as a concatenation of its lemma and PoS tag (for example, `\textit{love\_VERB}' and `\textit{love\_NOUN}' yield different word types). The mappings between BNC tags and Universal tags were created by us and released online\footnote{\url{http://bit.ly/291BlpZ}}.

The main motivation for the use of the Universal PoS tag set was that this is a newly emerging standard which is actively being used for annotation of a range of different languages through the community-driven Universal Dependencies effort \cite{nivre2016universal}. Additionally, this tag set is coarser than the original BNC one: it simplifies the workflow and eliminates the necessity to merge `inflectional' tags into one (e.g., singular and plural nouns into one `noun' class). This conforms with our interest in parts of speech proper, not inflectional forms within one PoS. We worked with the following 16 Universal tags: \textbf{ADJ, ADP, ADV, AUX, CONJ, DET, INTJ, NOUN, NUM, PART, PRON, PROPN, SCONJ, SYM, VERB, X} (punctuation tokens marked with the PUNCT tag were excluded).

Then, a \textit{Continuous Skipgram} embedding model \cite{mikolov2013distributed} was trained on this corpus, using a vector size of 300, 10 negative samples, a symmetric window of 2 words, no down-sampling, and 5 iterations over the training data. Words with corpus frequency less than 5 were ignored. This model represents the semantics of the words it contains. But at the same time, for each word, a PoS tag is known (from the BNC annotation). It means that is is possible to test how good the word embeddings are in grouping words according to their parts of speech.

To this end, we extracted vectors for the 10~000 most frequent words from the resulting model (roughly, these are the words with corpus frequency more than 500). Then, these vectors were used to train a simple logistic regression multinomial classifier aimed to predict the word's part of speech.

It is important that we applied classification, not clustering here. Naive \textit{K-Means} clustering of word embeddings in our model into 16 groups showed very poor performance (adjusted Rand index of 0.52 and adjusted Mutual Information score of 0.61 in comparison to the original BNC tags).
This is because PoS-related features form only a part of embeddings, and in the fully unsupervised setting, the words tend to cluster into semantic groups rather than `syntactic' ones. But when we train a classifier, it locates exactly the features (or combinations of features) that correspond to parts of speech, and uses them subsequently.

Note that during training (and subsequent testing), each word's vector was used several times, proportional to frequency of the word in the corpus, so the classifier was trained on 177~343 (sometimes repeating) instances, instead of the original 10~000. This was done to alleviate classification bias due to class imbalance: there are much fewer word types in the closed PoS classes (pronouns, conjunctions, etc.) than in the open ones (nouns, verbs, etc.), so without considering word frequency, the model does not have a chance to learn good predictors for `small' classes and ends up never predicting them. At the same time, words from closed classes occur very frequently in the running text, so after `weighting' training instances by corpus frequency, the balance is restored and the classifier model has enough training instances to learn to predict closed PoS classes as well. As an additional benefit, by this modification we make frequent words from all classes more `influential' in training the classifier. 

The resulting classifier showed a weighted macro-averaged F-score (over all PoS classes) and accuracy equal to 0.98, with 10-fold cross-validation on the training set. 
%
This is a significant improvement over the 
\textit{one-feature} baseline classifier (classify using only one vector dimension with maximum F-value in relation to class tags), with F-score equal to only 0.22. Thus, the results support the hypothesis that word embeddings contain information that allows us to group words together based on their parts of speech. At the same time, we see that this information is not restricted to some particular vector component: rather, it is distributed among several axis of the vector space.

After training the classifier, we were able to use it to detect `outlying' words in the BNC (judging by the distributional model). So as not to experiment on the same data we had trained our classifier on, we compiled another test set of 17 000 vectors for words with the BNC frequencies between 100 and 500. They were weighted by word frequencies in the same way as the training set, and the resulting test set contained 30 710 instances. Compared to the training error reported above we naturally observe a drop in performance when predicting PoS for this unseen data, but the classifier still appears quite robust, yielding an F-score of 0.91. However, some of the drop is also due to the fact that we are applying the classifier to words with lower frequency, and hence we have somewhat less training data for the input embeddings.

Furthermore, to make sure that the results can potentially be extended to other texts, we applied the trained classifier to all lemmas from the human-annotated Universal Dependencies English Treebank \cite{silveira14gold}. The words not present in the distributional model were omitted (they sum to 27\% of word types and 10\% of word tokens). The classifier showed an F-Score equal to 0.99, further demonstrating the robustness of the classifier. Note, however, that part of this performance is because the UD Treebank contains many words from the classifier training set. Essentially, it means that the decisions of the UD human annotators are highly consistent with the distributional patterns of words in the BNC.
%

In sum, the vast majority of words are classified correctly, which means that their embeddings enable the detection of their parts of speech. In fact, one can visualize `centroid' vectors for each PoS by simply averaging vectors of words belonging to this part of speech. We did this for 10 000 words from our training set. 

\begin{figure}
\begin{center}
\caption{Centroid embedding for coordinating conjunctions}\label{conj}
\includegraphics[scale=0.3,keepaspectratio]{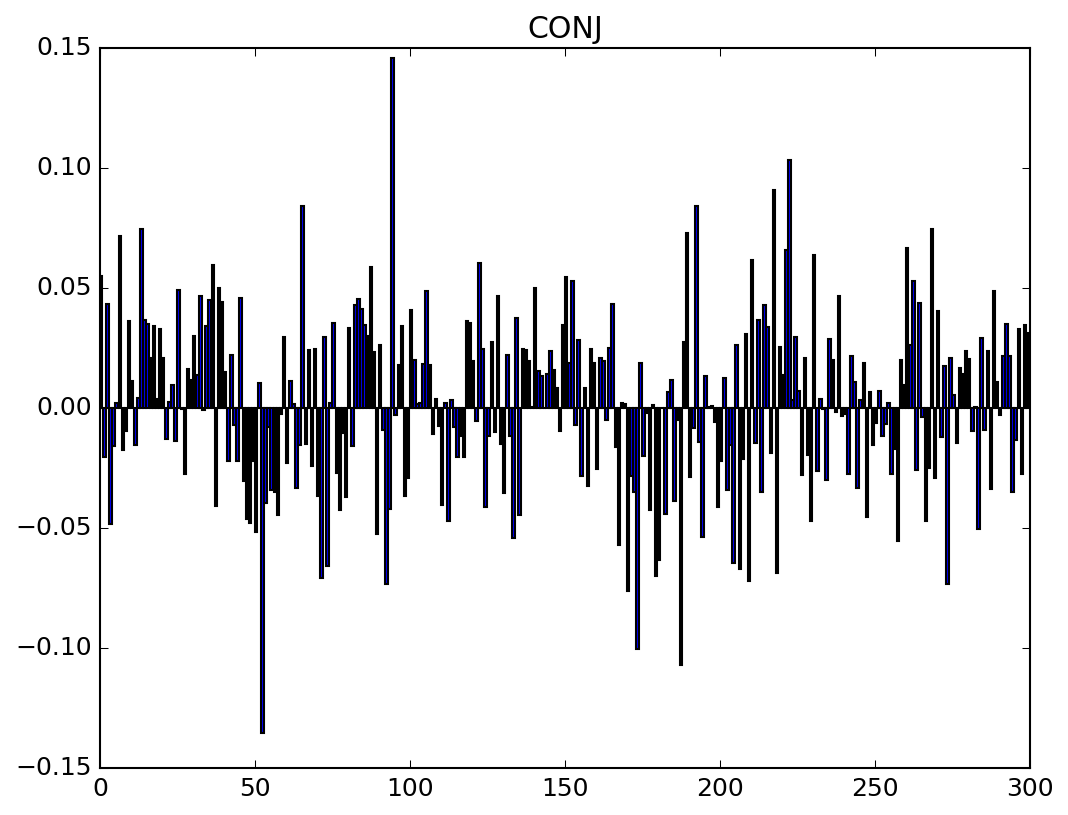}
\end{center}
\end{figure}

\begin{figure}
\begin{center}
\caption{Centroid embedding for subordinating conjunctions}\label{sconj}
\includegraphics[scale=0.3,keepaspectratio]{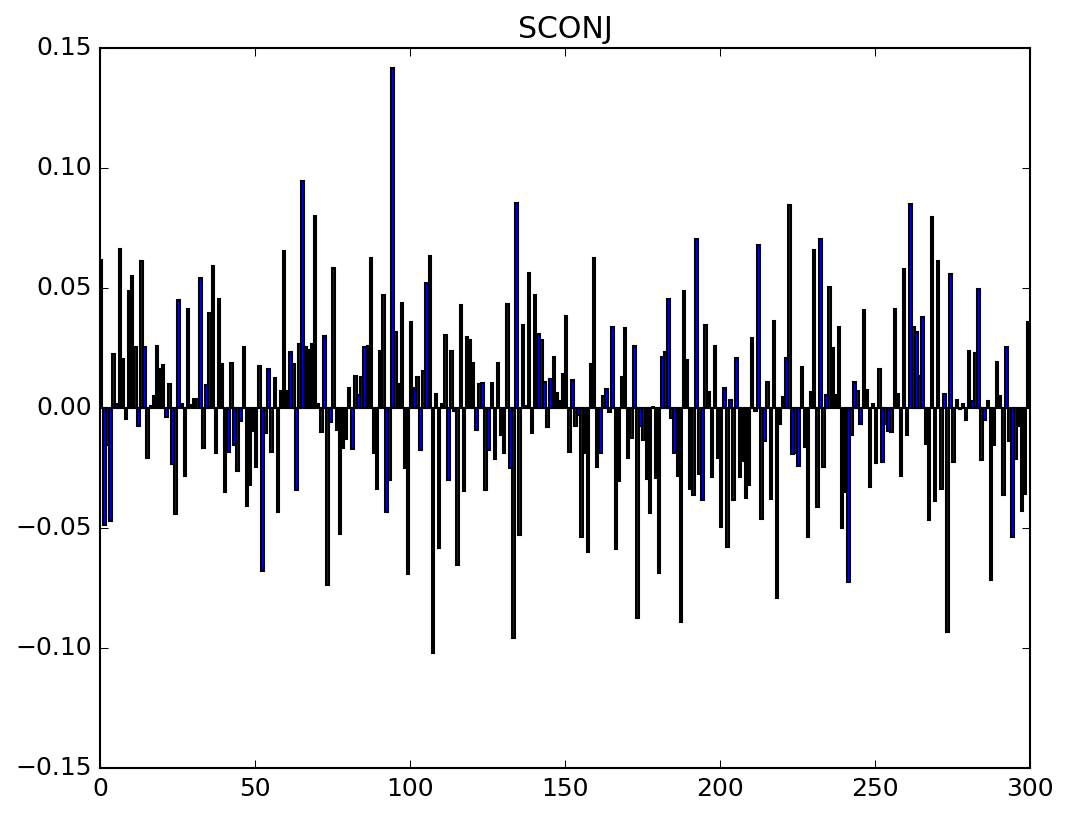}
\end{center}
\end{figure}

Plots for centroid vectors of coordinating and subordinating conjunctions are shown in Figures \ref{conj} and \ref{sconj} respectively. Even visually one can notice a very strongly expressed feature near the `100' mark in the horizontal axis (component number 94). In fact, this is indeed an idiosyncratic feature of conjunctions: none of the other parts of speech shows such a property.  More details about what vector components are relevant to part of speech affiliation are given in Section \ref{sec:predict}.

Additionally, with centroid PoS vectors we can find out how similar different parts of speech are to each other, by simply measuring cosine similarity between them. If we rank PoS pairs according to their similarity (Table \ref{tab:similar}), what we see is that nominative parts of speech are close to each other, determiners and pronouns are also similar, as well as prepositions and subordinating conjunctions; quite in accordance with linguistic intuition. Proper nouns are not very similar to common nouns, with cosine similarity between them only 0.67 (even adverbs are closer). Arguably, this is explained by co-occurrences together with the definite article, and as we show below, this helps the model to successfully separate the former from the latter.

\begin{table}
 \caption{Distributional similarity between parts of speech (fragment)}\label{tab:similar}
 \centering
\begin{tabular}{cc}
\toprule
Cosine similarity&PoS pair\\  \midrule
0.81&NOUN ADJ   \\   
0.77&ADV PRON  \\ 
0.73&DET PRON   \\
0.73&ADV ADJ   \\
...&...  \\
...&... \\
0.37&INTJ NUM   \\
0.36&AUX NUM  \\
\midrule
\end{tabular}
\end{table}
%

Despite generally good performance of the classifier, if we look at our BNC test set, 1741 word types (about 10\% of the whole test set vocabulary) were still classified incorrectly. Thus, they are somehow dissimilar to `prototypical' words of their parts of speech. These are the `outliers' we were after. We analyze the patterns found among them in the next section.

\section{Not from this crowd: analyzing outliers}\label{sec:crowd}
First, we filtered out misclassified word types with `\textbf{X}' BNC annotation (they are mostly foreign words or typos). This leaves us with 1558 words for which the classifier assigned part of speech tags different from the ones in the BNC. It probably means that these words' distributional patterns differ somehow from what is more typically observed, and that they tend to exhibit behavior similar to another part of speech. Table \ref{tab:rank} shows the most frequent misclassification cases, together accounting for more than 85\% of errors.
\begin{table}
 \caption{Most frequent PoS misclassifications of the distributional predictor. The \# column lists the number of word types.}\label{tab:rank}
 \centering
\begin{tabular}{@{}r@{\hskip 25px}l@{\hskip 15px}l@{}}
\toprule
\#&Actual PoS & Predicted PoS\\  \midrule
347& PROPN & NOUN \\   
313& ADJ & NOUN  \\ 
190& NOUN  & ADJ  \\
91& NOUN & PROPN  \\
87& PROPN & ADJ \\
57& VERB & ADJ \\
55& NOUN  & NUM  \\
52& NUM & NOUN  \\
45& NUM & PROPN \\
28& ADV & PROPN \\
25& ADV & NOUN \\
25& ADJ & PROPN \\
20& ADV & ADJ \\
\midrule
\end{tabular}
\end{table}

Additionally, we ranked misclassification cases by `part of speech coverage', that is by the ratio of the words belonging to a particular PoS for which our classifier outputs this particular type of misclassification. For example, proper nouns misclassified as common nouns constitute the most numerous error type in Table \ref{tab:rank}, but in fact only 9\% of all proper nouns in the test set were misclassified in this way. There are parts of speech with a much larger portion of word-types predicted erroneously: e.g., 22\% of subordinate conjunctions were classified as adverbs. Table \ref{tab:ratio} lists error types with the highest coverage (we excluded error types with absolute frequency equal to 1, as it is impossible to speculate on solitary cases).

We now describe some of the interesting cases. Almost 30\% of error types (judging by absolute amount of misclassified words) consist of proper nouns predicted to be common ones and vice versa. These cases do not tell us anything new, as it is obvious that distributionally these two classes of words are very similar, take the same syntactic contexts and hardly can be considered different parts of speech at all. At the same time, it is interesting that the majority of proper nouns in the test set (88\%) was correctly predicted as such. It means that in spite of contextual similarity, the distributional model has managed to extract features typical for proper names. Errors mostly cover comparatively rare names, such as `\textit{luftwaffe}', `\textit{stasi}', `\textit{stonehenge}', or `\textit{himalayas}'. Our guess is that the model was just not presented with enough contexts for these words to learn meaningful representations. Also, they are mostly not personal names but toponyms or organization names, probably occurring together with the definite article \textit{the}, unlike personal names. 

Another 30\% of errors are due to vague boundaries between nominal and adjectival distribution patterns in English: nouns can be modified by both (it seems that cases where a proper noun is mistaken for an adjective are often caused by the same factor). Words like `\textit{materialist\_NOUN}', `\textit{starboard\_NOUN}' or `\textit{hypertext\_NOUN}' are tagged as nouns in the BNC, but they often modify other nouns, and their contexts are so `adjectival' that the distributional model actually assigned them semantic features highly similar to those of adjectives. Vice versa, `\textit{white-collar\_ADJ}' (an adjective in BNC) is regarded as a noun from the point of view of our model. Indeed, there can be contradicting views on the correct part of speech for this word in phrases like `\textit{and all the other white-collar workers}'. Thus, in this case the distributional model highlights the already known similarity between two word classes. 

The cases of verbs mistaken for adjectives seem to be caused mostly by passive participles (`\textit{was overgrown}', `\textit{is indented}'), which intuitively are indeed very adjective-like. So, this gives us a set of verbs dominantly (or almost exclusively, like `\textit{to intertwine}' or `\textit{to disillusion}') used in passive. Of course, we will hardly announce such verbs to be adjectives based on that evidence, but at least we can be sure that this sub-class of verbs is clearly semantically and distributionally different from other verbs.

The next numerous type of errors consists of common nouns predicted to be numerals. A quick glance at the data reveals that 90\% of these `nouns' are in fact currency amounts and percentages (`\textit{£70}', `\textit{33\%}', `\textit{\$1}', etc). It seems reasonable to classify these as numerals, even though they contain some kind of nominative entities inside. Judging by the decisions of the classifier, their contexts do not differ much from those of simple numbers, and their semantics is similar. The Universal Dependencies Treebank is more consistent in this respect: it separates entities like `\textit{1\$}' into two tokens: a numeral (NUM) and a symbol (SYM). Consequently, when our classifier was tested on the words from the UD Treebank, there was only one occurrence of this type of error.

Related to this is the inverse case of numerals predicted to be common or proper nouns. It is interesting that this error type also ranks quite high in terms of coverage: if we combine numerals predicted to be common and proper nouns, we will see that 17\% of all numerals in the test set were subject to this error. The majority of these `numerals' are years (`\textit{1804}', `\textit{1776}', `\textit{1822}') and decades (`\textit{1820s}', `\textit{60s}' and even `\textit{twelfths}'). Intuitively, such entities do indeed function as nouns (`\textit{I'd like to return to the sixties}'). Anyway, it is difficult to invent a persuasive reason for why `\textit{fifty pounds}' should be tagged as a noun, but `\textit{the year 1776}' as a numeral. So, this points to possible (minor) inconsistencies in the annotation strategy of the BNC. Note that a similar problem exists in the Penn Treebank as well \cite{manning2011part}.

Adverbs classified as nouns (53 words in total for both common and proper nouns) are possibly the ones often followed by verbs or appearing in company of adjectives (examples are `\textit{ultra}' and `\textit{kinda}'). This made the model treat them as close to the nominative classes. Interestingly, most `adverbs' predicted to be proper nouns are time indicators (`\textit{7pm}', `\textit{11am}'); this also raises questions about what adverbial features are really present in these entities. Once again, unlike the BNC, the UD Treebank does not tag them as adverbs.

\begin{table}
 \caption{Coverage of misclassifications with distributional predictor, i.e., ratio of errors over all word types of a given PoS. The absolute type count is given by \#.}\label{tab:ratio}
 \centering
\begin{tabular}{@{}c@{\hskip 15px}l@{\hskip 15px}l@{\hskip 10px}r@{}}
\toprule
Coverage&Actual PoS & Predicted PoS & \# \\  \midrule
0.22& SCONJ & ADV & 2\\   
0.17& INTJ & PROPN & 8\\ 
0.11& ADP  & ADJ & 3 \\
0.09& ADJ & NOUN & 313 \\
0.09& PROPN & NOUN & 347\\
0.09& NUM & NOUN & 52\\
0.08& NUM  & PROPN & 45 \\
\midrule
\end{tabular}
\end{table}

The cases we described above revealed some inconsistencies in the BNC annotation. However, it seems that with adverbs mistaken for adjectives, we actually found a systematic error in the BNC tagging: these cases are mostly connected to adjectives like `\textit{plain}', `\textit{clear}' or `\textit{sharp}' (including comparative and superlative forms) erroneously tagged in the corpus as adverbs. These cases are not rare: just the three adjectives we mentioned alone appear in the BNC about 600 times with an adverb tag, mostly in clauses of the kind `\textit{the author makes it plain that\ldots}', so-called small clauses \cite{aarts2012small}. Sometimes these tokens are tagged as ambiguous, and the adjective tag is there as a second variant; however, the corpus documentation states that in such cases the first variant is always more likely. Thus, distributional models can actually detect outright errors in PoS-tagged corpora, when incorrectly tagged words strongly tend to cluster with another part of speech. In the UD treebank such examples can also be observed, but they are much fewer and more `adverbial', like `\textit{it goes \textbf{clear} through}'.

Turning to Table \ref{tab:ratio}, most of the entries were already covered above, except the first 3 cases. These relate to closed word classes (functional words), which is why the absolute number of influenced word types is low, but the coverage (ratio of all words of this PoS) is quite high.

First, out of 9 distinct subordinate conjunctions in the test set, 2 were predicted to be adverbs. This is not surprising, as these words are `\textit{seeing}' and `\textit{immediately}'. For `seeing' the prediction seems to be just a random guess (the prediction confidence was as low as 0.3), but with `\textit{immediately}' the classifier was actually more correct than the BNC tagger (the prediction confidence was about 0.5). In BNC, these words are mostly tagged as subordinate conjunctions in cases when they occur sentence-initially (`\textit{\textbf{Immediately}, she lowered the gun}'). The other words marked as SCONJ in the test set are really such, and the classifier made correct predictions matching the BNC tags.

Interjections mistaken for proper names do not seem very interpretable (examples are `\textit{gee}', `\textit{oy}' and `\textit{farewell}'). At the same time, 3 prepositions predicted to be adjectives clearly form a separate group: they are `\textit{cross}', `\textit{pre}' and `\textit{pro}'. They are not often used as separate words, but when they are (`\textit{Did anyone encounter any trouble from Hibs fans in Edinburgh \textbf{pre} season?}'), they are very close to adjectives or adverbs, so the predictions of the distributional classifier once again suggest shifting parts of speech boundaries a bit.

Error analysis on the vocabulary from the Universal Dependencies Treebank showed pretty much the same results, except for some differences already mentioned above.

There exists another way to retrieve this kind of data: to process tagged data with a conventional PoS tagger and analyze the resulting confusion matrix. We tested this approach by processing the whole BNC with the Stanford PoS Tagger \cite{toutanova2003feature}. Note that as an input to the tagger we used not the whole sentences from the corpora, but separate tokens, to mimic our workflow with the distributional predictor. Prior to this, BNC tags were converted to the Penn Treebank tagset\footnote{\url{https://www.cis.upenn.edu/~treebank/}} to match the output of the tagger. As we are interested in coarse, `overarching' word classes, inflectional forms were merged into one tag. 
That was easy to accomplish by dropping all characters of the tags after the first two (excluding proper noun tags, which were all converted to NNP). 

Analysis of the confusion matrix (cases where the tag predicted by the Stanford tagger was different from the BNC tag) revealed the most frequent error types shown in Table \ref{tab:stanfordrank}. Despite similar top positions of errors types `\textit{proper noun predicted as common noun}' and `\textit{nouns and adjectives mistaken for each other}', there are also very frequent errors of types `\textit{verb to noun}' and `\textit{adjective to verb}', not observed in the distributional confusion matrix (Table \ref{tab:rank}). We would not be able to draw the same insights that we did from the distributional confusion matrix: the case with verbs mistaken for adjective is ranked only 12th, adverbs mistaken for nouns - 13th, etc.

\begin{table}
 \caption{Most frequent PoS misclassifications with the Stanford tagger (counting word types).}\label{tab:stanfordrank}
 \centering
\begin{tabular}{@{}r@{\hskip 25px}l@{\hskip 20px}l@{}}
\toprule
\multicolumn{1}{c}{\#}&Actual & Predicted \\  \midrule
172675& NNP & NN \\   
47202& VB & NN  \\ 
40218& JJ  & NN  \\
24075& NN & JJ  \\
9723& JJ & VB \\
\midrule
\end{tabular}
\end{table}

Table \ref{tab:stanfordratio} shows top misclassification types by their word type coverage. Once again, interesting cases we discovered with the distributional confusion matrix (like subordinating conjunctions mistaken for adverbs and prepositions mistaken for adjectives) did not show up. Obviously, a lot of other insights can be extracted from the Stanford Tagger errors (as has been shown in previous work), but it seems that employing a distributional predictor reveals different error cases and thus is useful in evaluating the sanity of tag sets.

\begin{table}
 \caption{Coverage of misclassifications (from all word types of this PoS) with the Stanford tagger.}\label{tab:stanfordratio}
 \centering
\begin{tabular}{@{}c@{\hskip 15px}l@{\hskip 15px}l@{\hskip 10px}r@{}}
\toprule
Coverage&Actual & Predicted & \multicolumn{1}{c}{\#}\\ \midrule
0.91& NNP & NN & 172675\\   
0.8& UH & NN & 576\\ 
0.79& DT  & NN & 217 \\
0.78& EX & JJ & 11 \\
0.78& PR & NN & 517\\
\midrule
\end{tabular}
\end{table}

To sum up, the analysis of `boundary cases' detected by a classifier trained on distributional vectors, indeed reveals sub-classes of words lying on the verge between different parts of speech. It also allows for quickly discovering systematic errors or inconsistencies in PoS annotations, whether they be automatic or manual. Thus, discussions about PoS boundaries would benefit from taking this kind of data into consideration.

\section{Embeddings as PoS predictors}\label{sec:predict}
In the experiment described in the previous section, we used a model trained on words concatenated with their PoS tags. Thus, our `classifier' was a bit artificial in that it required a word plus a tag as an input, and then its output is a judgment about what tag is most applicable to this combination from the point of view of the BNC distributional patterns. This was not a problem for us, as our aim was exactly to discover lexical outliers.

But is it possible to construct a proper predictor in the same way, which is able to predict a PoS tag for a word without any pre-existing tags as hints? Preliminary experiments seem to indicate that it is.

We trained a \textit{Continuous Skipgram} distributional model on the BNC lemmas without PoS tags. After that, we constructed a vocabulary of all unambiguous lemmas from the UD Treebank training set. `Unambiguous' here means that the lemma either was always tagged with one and the same PoS tag in the Treebank, or has one `dominant' tag, with frequencies of other PoS assignments not exceeding 1/2 of the dominant assignment frequency. Our hypothesis was that these words are prototypical examples of their PoS classes, with corresponding prototypical features most pronounced; this approach is conceptually similar to \cite{haghighi2006prototype}. We also removed words with frequency less than 10 in the Treebank. This left us with 1564 words from all Universal Tag classes (excluding PUNCT, X and SYM, as we hardly want to predict punctuation or symbol tag).

Then the same simple logistic regression classifier was trained on the distributional vectors from the model for these 1564 words only, using UD Treebank tags as class labels (the training instances were again weighted proportionally to the words' frequencies in the Treebank). The resulting classifier showed an accuracy of 0.938 after 10-fold cross-validation on the training set. 

We then evaluated the classifier on tokens from the UD Treebank test set. Now the input to the classifier consisted of these tokens' lemmas only. Lemmas which were missing from the model's vocabulary were omitted (860 of a total of 21759 tokens in the test set). The model reached an accuracy of 0.84 (weighted precision 0.85, weighted recall 0.84).

These numbers may not seem very impressive in comparison with the performance of current state-of-the-art PoS taggers. However, one should remember that this classifier knows absolutely nothing about a word's context in the current sentence. It assigns PoS tags based solely on the proximity of the word's distributional vector in an unsupervised model to those of prototypical PoS examples. The classifier was in fact based only on knowledge of what words occurred in the BNC near other words within a symmetric window of 2 words to the left and to the right. It did not even have access to the information about exact word order within this sliding window, which makes its performance even more impressive.

It is also interesting that one needs as few as a thousand example words to train a decent classifier. Thus, it seems that PoS affiliation is expressed quite strongly and robustly in word embeddings. It can be employed, for example, in preliminary tagging of large corpora of resource-poor languages. Only a handful of non-ambiguous words need to be manually PoS-tagged, and the rest is done by a distributional model trained on the corpus.

Note that applying a \textit{K-neighbors} classifier instead of logistic regression returned somewhat lower results, with 0.913 accuracy on 10-fold cross-validation with the training set, and 0.81 accuracy on the test set. This seems to support our hypothesis that several particular embedding components correspond to part of speech affiliation, but not all of them. As a result, \textit{K-neighbors} classifier fails to separate these important features from all the others and predicts word class based on its nearest neighbors with all dimensions of the semantic space equally important. At the same time, logistic regression learns to pay more attention to the relevant features, neglecting unimportant ones.  

To find out how many features are important for the classifier, we used the same training and test set, and ranked all embedding components (features, vector dimensions) by their ANOVA F-value related to PoS class. Then we successively trained the classifier on increasing amounts of top-ranked features (top $k$ best) and measured the training set accuracy.

\begin{figure}
\caption{Classifier accuracy depending on the number of used vector components ($k$)}\label{fig:featurerank}
\includegraphics[width=\linewidth]{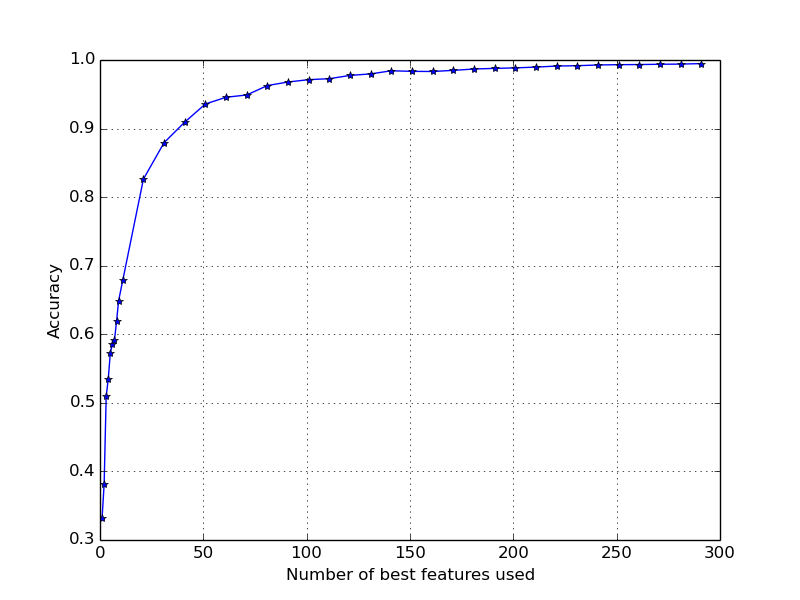}
\end{figure}

The results are shown in Figure \ref{fig:featurerank}. One can see that the accuracy smoothly grows with the number of used features, eventually reaching almost ideal performance on the training set. It is difficult to define the point where the influence of adding features reaches a plateau; it may lie somewhere near $k=100$. It means that the knowledge about PoS affiliation is distributed among at least one hundred components of the word embeddings, quite consistent with the underlying idea of embedding models.

One might argue that the largest gap in performance is between $k=2$ and $k=3$ (from 0.38 to 0.51) and thus most PoS-related information is contained in the 3 components with the largest F-value (in our case, these 3 features were components 31, 51 and 11). But an accuracy of 0.51 is certainly not an adequate result, so even if important, these components are not sufficient to robustly predict part of speech affiliation for a word. Further research is needed to study the effects of adding features to the classifier training.

Regardless, an interesting finding is that part of speech affiliation is distributed among many components of the word embeddings, not concentrated in one or two specific features. Thus, the strongly expressed component 94 in the average vector of conjunctions (Figures \ref{conj} and \ref{sconj}) seems to be a solitary case.  

\section{Conclusion}\label{sec:conclusion}
Distributional semantic vectors trained on word contexts from large text corpora can learn knowledge about part of speech clusters. Arguably, they are good at this precisely because part of speech boundaries are not strict, and even sometimes considered to be a non-categorical linguistic phenomenon \cite{manning2015computational}.

In this paper we have demonstrated that semantic features derived in the process of training a PoS prediction model on word embeddings can be employed both in supporting linguistic hypotheses about part of speech class changes and in detecting and fixing possible annotation errors in corpora. The prediction model is based on simple logistic regression and the word embeddings are trained using \textit{Continuous Skip-Gram} model over PoS-tagged lemmas. We show that the word embeddings contain robust data about the PoS classes of the corresponding words, and that this knowledge seems to be distributed among several components (at least a hundred in our case of 300-dimensional model). We also report preliminary results for predicting PoS tags using a classifier trained on a small number of prototypical members (words with a dominant PoS class) and applying it to embeddings estimated from unlabeled data. A detailed error analysis and experimental results are reported for both the BNC and the UD Treebank.

The reported experiment form part of ongoing research, and we plan to extend it, particularly conducting similar experiments with other languages typologically different from English.  We also plan to continue studying the issue of correspondence between particular embedding components and part of speech affiliation. Another direction of future work is finding out how different hyperparameters for training distributional models (including training corpus pre-processing) influence their performance in PoS discrimination, and also comparing the results to using structured embedding models like those of \newcite{ling2015twotoo}.

\bibliographystyle{acl2016}
\bibliography{conll}

\end{document}